# Pattern Recognition for Conditionally Independent Data

Daniil Ryabko
daniil@ryabko.net
IDSIA, Galleria 2, 6928 Manno-Lugano Switzerland;
Computer Learning Research Centre
Royal Holloway, University of London

September 25, 2018

## Abstract

[1] In this work we consider the task of relaxing the i.i.d assumption in pattern recognition (or classification), aiming to make existing learning algorithms applicable to a wider range of tasks. Pattern recognition is guessing a discrete label of some object based on a set of given examples (pairs of objects and labels). We consider the case of deterministically defined labels. Traditionally, this task is studied under the assumption that examples are independent and identically distributed. However, it turns out that many results of pattern recognition theory carry over a weaker assumption. Namely, under the assumption of conditional independence and identical distribution of objects, while the only assumption on the distribution of labels is that the rate of occurrence of each label should be above some positive threshold.

We find a broad class of learning algorithms for which estimations of the probability of a classification error achieved under the classical i.i.d. assumption can be generalised to the similar estimates for the case of conditionally i.i.d. examples.

---

# 1 Introduction

Pattern recognition (or classification) is, informally, the following task. There is a finite number of classes of some complex objects. A predictor is learning to label objects according to the class they belong to (i.e. to classify), based only on some examples (labelled objects). One of the typical practical examples is recognition of a hand-written text. In this case, an object is a hand-written letter and a label is the letter of an alphabet it represents. Other examples include DNA sequence identification, recognition of an illness based on a set of symptoms, speech recognition, and many others.

The formal model of the task used most widely is described, for example, in [28], and can be briefly introduced as follows (we will later refer to it as "the i.i.d. model"). The objects $x \in \mathbf{X}$ are drawn independently and identically distributed (i.i.d.) according to some unknown (but fixed) probability distribution $P(x)$. The labels $y \in \mathbf{Y}$ are given for each object according to some (also unknown but fixed) function[2] $\eta(x)$. The space $\mathbf{Y}$ of labels is assumed to be finite (often binary). The task is to construct the best predictor for the labels, based on the data observed, i.e. actually to "learn" $\eta(x)$.

This task is usually considered in either of the following two settings. In off-line setting a (finite) set of examples is divided into two finite subsets, the training set and the testing set. A predictor is constructed based on the first set and then is used to classify the objects from the second. In online setting a predictor starts by classifying the first object with zero knowledge; then it is given the correct label and (having "learned" this information) proceeds with classifying the second object, the correct second label is given, and so on.

There is a plenty of algorithms developed for solving pattern recognition tasks (see [10, 28, 15] for the most widely used methods). However, the i.i.d assumption, which is central in the model, is too tight for many applications. It turns out that it is also too tight for a wide range of methods developed under the assumptions of the model: they work nearly as well under weaker conditions.

First consider the following example. Suppose we are trying to recognise a hand-written text. Obviously, letters in the text are dependent (for example, we strongly expect to meet "u" after "q"). This seemingly implies that pattern recognition can not be applied to this task, which is, however, one of their classical applications.

---
[2]Often (e.g. in [28]) a more general situation is considered, the labels are drawn according to some probability distribution $P(y|x)$, i.e. each object can have more than one possible label.

We show that the following two assumptions on the distribution of examples are sufficient for pattern recognition. First, that the dependence between objects is only that between their labels and the type of object-label dependence does not change in time.

These intuitive ideas lead us to the following model (to which we refer as "the conditional model"). The labels $y \in \mathbf{Y}$ are drawn according to some unknown (but fixed) distribution over the set of all infinite sequences of labels. There can be any type of dependence between labels; moreover, we can assume that we are dealing with any (fixed) combinatorial sequence of labels. However, in this sequence the rate of occurrence of each label should keep above some positive threshold. For each label $y$ the corresponding object $x \in \mathbf{X}$ is generated according to some (unknown but fixed) probability distribution $P(x|y)$. All the rest is as in the i.i.d. model.

The main difference from the i.i.d. model is in that in the conditional model we made the distribution of labels primal; having done that we can relax the requirement of independence of objects to the conditional independence.

In this work we provide a tool for obtaining estimations of probability of error of a predictor in the conditional model from an estimation of the probability of error in the i.i.d. model. The general theorems about extending results concerning performance of a predictor to the conditional model are illustrated on two classes of predictors.

First, we extend weak consistency results concerning partitioning and nearest neighbour estimates from the i.i.d. model to the conditional model.

Second, we use some results of Vapnik-Chervonenkis theory to estimate performance in the conditional model (on finite amount of data) of predictors minimising empirical risk, and also obtain some strong consistency results.

These results are obtained as applications of the following rule. The only assumption on a predictor under which a predictor works in the new model as well as in the i.i.d. model is what we call *tolerance to data*: in any large dataset there is no small subset which strongly changes the probability of error. This property should also hold with respect to permutations. This assumption on a predictor should be valid in the i.i.d. model. Thus, the results achieved in the i.i.d. model can be extended to the conditional model; this concerns distribution–free results as well as distribution–specific, results on the performance on finite samples as well as asymptotic results.

Various approaches to relaxing the i.i.d. assumption in learning tasks have been proposed in the literature. Thus, in [17, 16] the authors study the nearest neighbour and kernel estimators for the task of regression estimation with continuous regression function, under the assumption that labels are conditionally independent given their objects, while objects form any individual sequence. Another approach is considered in [20], where a regression estimation scheme is proposed which is consistent for any individual stable sequence of object-label pairs (no probabilistic assumptions), assuming that there is a known upper bound on the variation of regression function.

There are also several approaches in which different types of assumptions on the joint distribution of objects and labels are made; then the authors construct a predictor or a class of predictors, to work well under the assumptions made. Thus, in [13] and [1] a generalisation of PAC approach to Markov chains with finite or countable state space is presented. The estimates of probability of error are constructed for this cases, under the assumption that the optimal rule generating examples belongs to a pre-specified class of decision rules. There is also a track of research on prediction under the assumption that the distribution generating examples is stationary and ergodic. The basic difference from our learning task, apart from different probabilistic assumption, is in that we are only concerned with object-label dependence, while in predicting ergodic sequences it is label-label (time series) dependence that is of primary interest. On this task see [24, 2, 21, 22] and references therein. Another approach is taken in [12, 3] where the PAC model is generalised to allow concepts changing over time. Here the methodology is proposed to track time series dependences, that is the authors find some classes of dependences which can be exploited for learning. Again the difference with our approach is that we try to find a (broad) class of problems where the time series dependence can be ignored by any reasonable pattern recognition method rather than constructing methods to use some specific dependences of this kind.

## 2  Definitions and General Results

Consider a sequence of *examples* $(x_1, y_1), (x_2, y_2), \ldots$; each example $z_i := (x_i, y_i)$ consists of an *object* $x_i \in \mathbf{X}$ and a *label* $y_i := \eta(x_i) \in \mathbf{Y}$, where $\mathbf{X}$ is a measurable space called an *object space*, $\mathbf{Y} := \{0, 1\}$ is called a *label space* and $\eta : \mathbf{X} \to \mathbf{Y}$ is some deterministic function. For simplicity we made the assumption that the space $\mathbf{Y}$ is binary, but all results easily extend to the case of any finite space $\mathbf{Y}$. The notation $\mathbf{Z} := \mathbf{X} \times \mathbf{Y}$ is used for the measurable space of examples. Objects are drawn according to some probability distribution $\mathbf{P}$ on $\mathbf{X}^\infty$ (and labels are defined by $\eta$). Thus we consider only the case of deterministically defined labels (that is, the noise-free model); in section 5 we discuss possible generalisations.

The notation $\mathbf{P}$ is used for distributions on $\mathbf{X}^\infty$ while the symbol $P$ is reserved for distributions on $\mathbf{X}$. In the latter case $P^\infty$ denotes the i.i.d. distribution on $\mathbf{X}^\infty$ generated by $P$. Correspondingly we will use symbols $\mathbf{E}$, $E$ and $\mathbf{E}^\infty$ for expectations over spaces $\mathbf{X}^\infty$ and $\mathbf{X}$. Letters $x, y, z$ (with indices) will be used for elements of spaces $\mathbf{X}, \mathbf{Y}, \mathbf{Z}$ correspondingly, while letters $X, Y, Z$ are reserved for random variables on these spaces.

The traditional assumption about the distribution $\mathbf{P}$ generating objects is that examples are independently and identically distributed (i.i.d.) according to some distribution $P$ on $\mathbf{X}$ (i.e. $\mathbf{P} = P^\infty$).

Here we replace this assumption with the following two conditions.

*First*, for any $n \in \mathbb{N}$ and for any measurable set $A \subset \mathbf{X}$

$$\mathbf{P}(X_n \in A \mid Y_n, X_1, Y_1, \ldots, X_{n-1}, Y_{n-1}) = \mathbf{P}(X_n \in A \mid Y_n) \qquad (1)$$

(i.e. some versions of conditional probabilities coincide). This condition looks very much like Markov condition which requires that each object depends on the past only through its immediate predecessor. The condition (1) says that each object depends on the past only through its label.

*Second*, for any $y \in \mathbf{Y}$, for any $n_1, n_2 \in \mathbb{N}$ and for any measurable set $A \subset \mathbf{X}$

$$\mathbf{P}(X_{n_1} \in A \mid Y_{n_1} = y) = \mathbf{P}(X_{n_2} \in A \mid Y_{n_2} = y) \qquad (2)$$

(i.e. the process is uniform in time; (1) allows dependence in $n$).

Note that the first condition means that objects are conditionally independent given labels (on conditional independence see [7]). Under the conditions (1) and (2) we say that *objects are conditionally independent and identically distributed* (conditionally i.i.d).

For each $y \in \mathbf{Y}$ denote the distribution $\mathbf{P}(X_n \mid Y_n = y)$ by $P_y$ (it does not depend on $n$ by (2) ). Clearly, the distributions $P_0$ and $P_1$ define some distributions $P$ on $\mathbf{X}$ up to a parameter $p \in [0, 1]$. That is, $P_p(A) = pP_1(A) + (1-p)P_0(A)$ for any measurable set $A \subset \mathbf{X}$ and for each $p \in [0, 1]$. Thus with each distribution $\mathbf{P}$ satisfying the assumptions (1) and (2) we will associate a family of distributions $P_p$, $p \in [0, 1]$.

The assumptions of the conditional model can be also interpreted as follows. Assume that we have some individual sequence $(y_n)_{n\in\mathbb{N}}$ of labels and two probability distributions $P_0$ and $P_1$ on $\mathbf{X}$, such that there exists sets $X_0$ and $X_1$ in $\mathbf{X}$ such that $P_1(X_1) = P_0(X_0) = 1$ and $P_0(X_1) = P_1(X_0) = 0$ (i.e. $X_0$ and $X_1$ define some function $\eta$). Each example $x_n \in \mathbf{X}$ is drawn according to the distribution $P_{y_n}$; examples are drawn independently of each other.

A *predictor* is a measurable function $\Gamma_n := \Gamma(x_1, y_1, \ldots, x_n, y_n, x_{n+1})$ taking values in $\mathbf{Y}$ (more formally, a family of functions indexed by $n$).

The probability of error of a predictor $\Gamma$ on each step $n$ is defined as

$$\mathrm{err}_n(\Gamma, \mathbf{P}, z_1, \ldots, z_n) := \mathbf{P}\big\{(x, y) \in \mathbf{Z} : y \neq \Gamma_n(z_1, \ldots, z_n, x)\big\}$$

($z_i$, $1 \leq i \leq n$ are fixed and the probability is taken over $z_{n+1}$). We will sometimes omit some of the arguments of $\mathrm{err}_n$ when it can cause no confusion; in particular, we will often use a short notation $\mathbf{P}(\mathrm{err}_n(\Gamma, Z_1, \ldots, Z_n) > \varepsilon)$ and an even shorter one $\mathbf{P}(\mathrm{err}_n(\Gamma) > \varepsilon)$ in place of

$$\mathbf{P}\big\{z_1, \ldots, z_n : \mathrm{err}_n(\Gamma, \mathbf{P}, z_1, \ldots, z_n) > \varepsilon\big\}.$$

For a pair of distributions $P_0$ and $P_1$ and any $\delta \in (0, 1/2)$ define

$$\triangledown_\delta(P_0, P_1, n, \varepsilon) := \sup_{p \in [\delta, 1-\delta]} P_p^\infty(\mathrm{err}_n(\Gamma) > \varepsilon) \tag{3}$$

For a predictor $\Gamma$ and a distribution $P$ on $\mathbf{X}$ define

$$\Delta(P, n, z_1, \ldots, z_n) := \max_{j \leq \varkappa_n;\ \pi:\{1,\ldots,n\}\to\{1,\ldots,n\}} \big| \mathrm{err}_n(\Gamma, P^\infty, z_1, \ldots, z_n) - \mathrm{err}_{n-j}(\Gamma, P^\infty, z_{\pi(1)}, \ldots, z_{\pi(n-j)}) \big|.$$

Define the *tolerance to data* of $\Gamma$ as

$$\Delta(P, n, \varepsilon) := P^n\big(\Delta(P, n, Z_1, \ldots, Z_n) > \varepsilon\big) \tag{4}$$

for any $n \in \mathbb{N}$, any $\varepsilon > 0$ and $\varkappa_n := \sqrt{n \log n}$ (see the end of Section 5 for the discussion of the choice of the constants $\varkappa_n$). Furthermore, for a pair of distributions $P_0$ and $P_1$ and any $\delta \in (0, 1/2)$ define

$$\Delta_\delta(P_0, P_1, n, \varepsilon) := \sup_{p \in [\delta, 1-\delta]} \Delta(P_p, n, \varepsilon).$$

Tolerance to data means, in effect, that in any typical large portion of data there is no small portion that changes strongly the probability of error. This property should also hold with respect to permutations.

We will also use another version of tolerance to data, in which instead of removing some examples we replace them with an arbitrary sample $z'_j, \ldots, z'_n$ consistent with $\eta$:

$$\bar{\Delta}(P, z_1, \ldots, z_n) := \sup_{j < \varkappa_n; \pi: \{1,\ldots,n\} \to \{1,\ldots,n\}; z'_{n-j}, \ldots, z'_n}$$
$$|\operatorname{err}_n(\Gamma, P^\infty, z_1, \ldots, z_n) - \operatorname{err}_n(\Gamma, P^\infty, \zeta_1, \ldots, \zeta_n)|,$$

where $\zeta_{\pi(i)} := z_{\pi(i)}$ if $i < n - j$ and $\zeta_{\pi(i)} := z'_i$ otherwise; the maximum is taken over all $z'_i$, $n - j < i \leq n$ consistent with $\eta$. Define

$$\bar{\Delta}(P, n, \varepsilon) := P^n\big(\bar{\Delta}(P, n, Z_1, \ldots, Z_n) > \varepsilon\big)$$

and

$$\bar{\Delta}_\delta(P_0, P_1, n, \varepsilon) := \sup_{p \in [\delta, 1-\delta]} \bar{\Delta}(P_p, n, \varepsilon).$$

The same notational convention will be applied to $\Delta$ and $\bar{\Delta}$ as to $\operatorname{err}_n$.

Various notions similar to tolerance to data have been studied in literature. Perhaps first they appeared in connection with deleted or condensed estimates (see e.g. [23]), and were later called stability (see [6, 14] for present studies of different kinds of stability, and for extensive overviews). Naturally, such notions arise when there is a need to study the behaviour of a predictor when some of the training examples are removed. These notions are much similar to what we call tolerance to data, only we are interested in the maximal deviation of probability of error while usually it is the average or minimal deviations that are estimated.

A predictor developed to work in the off-line setting should be, loosely speaking, tolerant to small changes in a training sample. The next theorem shows under which conditions this property of a predictor can be utilised.

**Theorem 1.** *Suppose that a distribution $\mathbf{P}$ generating examples is such that the objects are conditionally i.i.d, i.e. $\mathbf{P}$ satisfies (1) and (2). Fix some $\delta \in (0, 1/2]$, let $p(n) := \frac{1}{n}\#\{i \leq n : Y_i = 1\}$ and $C_n := \mathbf{P}(\delta \leq p(n) \leq 1 - \delta)$ for each $n \in \mathbb{N}$. Let also $\alpha_n := \frac{1}{1 - 1/\sqrt{n}}$. For any predictor $\Gamma$ and any $\varepsilon > 0$ we have*

$$\mathbf{P}(\operatorname{err}_n(\Gamma) > \varepsilon) \leq C_n^{-1} \alpha_n \big(\nabla_\delta(P_0, P_1, n + \varkappa_n, \delta\varepsilon/2) \\ + \Delta_\delta(P_0, P_1, n + \varkappa_n, \delta\varepsilon/2)\big) + (1 - C_n), \quad (5)$$

*and*

$$\mathbf{P}(\mathrm{err}_n(\Gamma) > \varepsilon) \leq C_n^{-1}\alpha_n\big(\nabla_\delta(P_0, P_1, n, \delta\varepsilon/2) \\ + \bar{\Delta}_\delta(P_0, P_1, n, \delta\varepsilon/2)\big) + (1 - C_n). \quad (6)$$

The proofs for this section can be found in Appendix A.

The theorem says that if we know with some confidence $C_n$ that the rate of occurrence of each label is not less than some (small) $\delta$, and have some bounds on the error rate and tolerance to data of a predictor in the i.i.d. model, then we can obtain bounds on its error rate in the conditional model.

Thus we have a tool for estimating the performance of a predictor on each finite step $n$. In Section 4 we will show how this result can be applied to predictors minimising empirical risk. However, if we are only interested in asymptotic results the formulations can be somewhat simplified.

Consider the following asymptotic condition on the frequencies of labels. Define $p(n) := \frac{1}{n}\#\{i \leq n : Y_i = 1\}$. We say that the *rates of occurrence of labels are bounded from below* if there exist such $\delta$, $0 < \delta < 1/2$ that

$$\lim_{n \to \infty} \mathbf{P}(p(n) \in [\delta, 1 - \delta]) = 1. \quad (7)$$

As the condition (7) means $C_n \to 1$ we can derive from Theorem 1 the following corollary.

**Corollary 1.** *Suppose that a distribution $\mathbf{P}$ satisfies (1), (2), and (7) for some $\delta \in (0, 1/2]$. Let $\Gamma$ be such a predictor that*

$$\lim_{n \to \infty} \nabla_\delta(P_0, P_1, n, \varepsilon) = 0 \quad (8)$$

*and either*

$$\lim_{n \to \infty} \Delta_\delta(P_0, P_1, n, \varepsilon) = 0 \quad (9)$$

*or*

$$\lim_{n \to \infty} \bar{\Delta}_\delta(P_0, P_1, n, \varepsilon) = 0 \quad (10)$$

*for any $\varepsilon > 0$. Then*

$$\mathbf{E}(\mathrm{err}_n(\Gamma, \mathbf{P}, Z_1, \ldots, Z_n)) \to 0.$$

In Section 3 we show how this statement can be applied to prove weak consistence of some classical nonparametric predictors in the conditional model.

# 3 Application to classical nonparametric predictors

In this section we will consider two types of classical nonparametric predictors: partitioning and nearest neighbour classifiers.

The nearest neighbour predictor assigns to a new object $x_{n+1}$ the label of its nearest neighbour among $x_1, \ldots, x_n$:

$$\Gamma_n(x_1, y_1, \ldots, x_n, y_n, x_{n+1}) := y_j,$$

where $j := \operatorname{argmin}_{i=1,\ldots,n} \|x - x_i\|$.

For i.i.d. distributions this predictor is also consistent, i.e.

$$E^\infty(\operatorname{err}_n(\Gamma, P^\infty)) \to 0,$$

for any distribution $P$ on $\mathbf{X}$ (see [8]).

We generalise this result as follows.

**Theorem 2.** *Let $\Gamma$ be the nearest neighbour classifier. Let $\mathbf{P}$ be some distribution on $\mathbf{X}^\infty$ satisfying (1), (2) and (7). Then*

$$\mathbf{E}(\operatorname{err}_n(\Gamma, \mathbf{P})) \to 0.$$

The proofs for this section can be found in Appendix B.

A partitioning predictor on each step $n$ partitions the object space $\mathbf{X} = \mathbb{R}^d$, $d \in \mathbb{N}$ into disjoint cells $A_1^n, A_2^n, \ldots$ and classifies in each cell according to the majority vote:

$$\Gamma(z_1, \ldots, z_n, x) := \begin{cases} 0 & \text{if } \sum_{i=1}^n I_{y_i=1} I_{x_i \in A(x)} \leq \sum_{i=1}^n I_{y_i=0} I_{x_i \in A(x)} \\ 1 & \text{otherwise,} \end{cases}$$

where $A(x)$ denotes the cell containing $x$. Define

$$\operatorname{diam}(A) := \sup_{x,y \in A} \|x - y\|$$

and

$$N(x) := \sum_{i=1}^n I_{x_i \in A(x)}.$$

It is a well known result (see, e.g. [10]) that a partitioning predictor is weakly consistent, provided certain regulatory conditions on the size of cells.

More precisely, let $\Gamma$ be a partitioning predictor such that $\mathrm{diam}(A(X)) \to 0$ in probability and $N(X) \to \infty$ in probability. Then for any distribution $P$ on $\mathbf{X}$
$$E^\infty(\mathrm{err}_n(\Gamma, P^\infty)) \to 0.$$

We generalise this result to the case of conditionally i.i.d. examples as follows.

**Theorem 3.** *Let $\Gamma$ be a partitioning predictor such that $\mathrm{diam}(A(X)) \to 0$ in probability and $N(X) \to \infty$ in probability, for any distribution generating i.i.d. examples. Then*
$$\mathbf{E}(\mathrm{err}_n(\Gamma, \mathbf{P})) \to 0$$
*for any distribution $\mathbf{P}$ on $\mathbf{X}^\infty$ satisfying (1), (2) and (7).*

Observe that we only generalise results concerning weak consistency of (one) nearest neighbour and non-data-dependent partitioning rules. More general results exist (see e.g. [9],[18]), in particular for data-dependent rules. However, we do not aim to generalise state-of-the-art results in nonparametric classification, but rather to illustrate that weak consistency results can be extended to the conditional model.

## 4 Application to Empirical Risk Minimisation.

In this section we show how to estimate the performance of a predictor minimising empirical risk (over certain class of functions) using Theorem 1. To do this we estimate the tolerance to data of such predictors, using some results from Vapnik-Chervonenkis theory. For the overviews of Vapnik-Chervonenkis theory see [29, 28, 10].

Let $\mathbf{X} = \mathbb{R}^d$ for some $d \in \mathbb{N}$ and let $\mathcal{C}$ be a class of measurable functions of the form $\varphi : \mathbf{X} \to \mathbf{Y} = \{0,1\}$, called *decision functions*. For a probability distribution $P$ on $\mathbf{X}$ define $\mathrm{err}(\varphi, P) := P(\varphi(X_i) \neq Y_i)$. If the examples are generated i.i.d. according to some distribution $P$, the aim is to find a function $\varphi$ from $\mathcal{C}$ for which $\mathrm{err}(\varphi, P)$ is minimal:

$$\varphi_P = \mathrm{argmin}_{\varphi \in \mathcal{C}}\, \mathrm{err}(\varphi, P).$$

In the theory of empirical risk minimisation this function is approximated by the function
$$\varphi_n^* := \arg\min_{\varphi \in \mathcal{C}} \overline{\mathrm{err}}_n(\varphi)$$
where $\overline{\mathrm{err}}_n(\varphi) := \sum_{i=1}^n I_{\varphi(X_i) \neq Y_i}$ is the empirical error functional, based on a sample $(X_i, Y_i)$, $i = 1, \ldots, n$. Thus, $\Gamma_n(z_1, \ldots, z_n, x_{n+1}) := \varphi_n^*(x_{n+1})$ is a predictor minimising empirical risk over the class of functions $\mathcal{C}$.

One of the basic results of Vapnik-Chervonenkis theory is the estimation of the difference of probabilities of error between the best possible function in the class $(\varphi_P)$ and the function which minimises empirical error:
$$P\big(\mathrm{err}_n(\Gamma, P^\infty) - \mathrm{err}(\varphi_P, P) > \varepsilon\big) \leq 8\mathcal{S}(\mathcal{C}, n)e^{-n\varepsilon^2/128},$$
where the symbol $\mathcal{S}(\mathcal{C}, n)$ is used for the $n$-th shatter coefficient of the class $\mathcal{C}$:
$$\mathcal{S}(\mathcal{C}, n) := \max_{A:=\{x_1,\ldots,x_n\} \subset \mathbf{X}} \#\{C \cap A : C \in \mathcal{C}\}.$$
Thus,
$$P(\mathrm{err}_n(\Gamma) > \varepsilon) \leq I_{\mathrm{err}(\varphi_P, P) > \varepsilon/2} + 8\mathcal{S}(\mathcal{C}, n)e^{-n\varepsilon^2/512}.$$
A particularly interesting case is when the optimal rule belongs to $\mathcal{C}$, i.e. when $\eta \in \mathcal{C}$. This situation was investigated in e.g. [27, 5]. Obviously, in this case $\varphi_P \in \mathcal{C}$ and $\mathrm{err}(\varphi_P, P) = 0$ for any $P$. Moreover, a better bound exists (see [28, 5, 10])
$$P(\mathrm{err}_n(\Gamma, P) > \varepsilon) \leq 2\mathcal{S}(\mathcal{C}, n)e^{-n\varepsilon/2}.$$

**Theorem 4.** *Let $\mathcal{C}$ be a class of decision functions and let $\Gamma$ be a predictor which for each $n \in \mathbb{N}$ minimises $\overline{\mathrm{err}}_n$ over $\mathcal{C}$ on the observed examples $(z_1, \ldots, z_n)$. Fix some $\delta \in (0, 1/2]$, let $p(n) := \frac{1}{n}\#\{i \leq n : Y_i = 0\}$ and $C_n := \mathbf{P}(\delta \leq p(n) \leq 1 - \delta)$ for each $n \in \mathbb{N}$. Assume $n > 4/\varepsilon^2$ and let $\alpha_n := \frac{1}{1 - 1/\sqrt{n}}$. We have*
$$\Delta(P_0, P_1, n, \varepsilon) \leq 16\mathcal{S}(\mathcal{C}, n)e^{-n\varepsilon^2/512}. \tag{11}$$
*(which does not depend on the distributions $P_0$ and $P_1$) and*
$$\mathbf{P}(\mathrm{err}_n(\Gamma, \mathbf{P}) > \varepsilon) \leq I_{2\,\mathrm{err}(\varphi_{P_{1/2}}, P_{1/2}) > \varepsilon/2} \tag{12}$$
$$+ 16\alpha_n C_n^{-1} \mathcal{S}(\mathcal{C}, n)e^{-n\delta^2\varepsilon^2/2048} + (1 - C_n).$$

*If in addition $\eta \in \mathcal{C}$ then*

$$\Delta(n,\varepsilon) \leq 4\mathcal{S}(\mathcal{C}, 2n)2^{-n\varepsilon/8} \qquad (13)$$

*and*

$$\mathbf{P}(\mathrm{err}_n(\Gamma, \mathbf{P}) > \varepsilon) \leq 4\alpha_n C_n^{-1}\mathcal{S}(\mathcal{C}, n)e^{-n\delta\varepsilon/16} + (1 - C_n). \qquad (14)$$

Thus, if we have bounds on the VC dimension of some class of classifiers, we can obtain bounds on the performance of predictors minimising empirical error for the conditional model.

Next we show how strong consistency results can be achieved in the conditional model. For general strong universal consistency results (with examples) see [19, 28, 29].

Denote the VC dimension of $\mathcal{C}$ by $V(\mathcal{C})$:

$$V(\mathcal{C}) := \max\{n \in \mathbb{N} : \mathcal{S}(\mathcal{C}, n) = 2^n\}.$$

Using Theorem 4 and Borel-Cantelli lemma, we obtain the following corollary.

**Corollary 2.** *Let $\mathcal{C}^k$, $k \in \mathbb{N}$ be a sequence of classes of decision functions with finite VC dimension such that $\lim_{k \to 0} \inf_{\varphi \in \mathcal{C}^k} \mathrm{err}(\varphi, P) = 0$ for any distribution $P$ on $\mathbf{X}$. If $k_n \to \infty$ and $\frac{V(\mathcal{C}^{k_n}) \log n}{n} \to 0$ as $n \to \infty$ then*

$$\mathrm{err}(\Gamma, \mathbf{P}) \to 0 \quad \mathbf{P}\text{-}a.s.$$

*where $\Gamma$ is a predictor which in each trial $n$ minimises empirical risk over $\mathcal{C}^{k_n}$ and $\mathbf{P}$ is any distribution satisfying (1), (2) and $\sum_{n=1}^{\infty}(1 - C_n) < \infty$.*

In particular, if we use bound on the VC dimension on classes of neural networks provided in [4] then we obtain the following corollary.

**Corollary 3.** *Let $\Gamma$ be a classifier that minimises the empirical error over the class $\mathcal{C}^{(k)}$, where $\mathcal{C}^{(k)}$ is the class of neural net classifiers with $k$ nodes in the hidden layer and the threshold sigmoid, and $k \to \infty$ so that $k \log n / n \to 0$ as $n \to \infty$. Let $\mathbf{P}$ be any distribution on $\mathbf{X}^{\infty}$ satisfying (1) and (2) such that $\sum_{n=1}^{\infty}(1 - C_n) < \infty$. Then*

$$\lim_{n \to \infty} \mathrm{err}_n(\Gamma) = 0 \quad \mathbf{P}\text{-}a.s.$$

# 5 Discussion

We have introduced "conditionally i.i.d." model for pattern recognition which generalises the commonly used i.i.d. model. Naturally, a question arises whether our conditions on the distributions and on predictors are necessary, or they can be yet more generalised in the same direction. In this section we discuss the conditions of the new model from this point of view.

The first question is, can the same results be obtained without assumptions on tolerance to data? The following negative example shows that some bounds on tolerance to data are necessary.

**Remark 1.** *There exists a distribution $\mathbf{P}$ on $\mathbf{X}^\infty$ satisfying (1) and (2) such that $\mathbf{P}(|p_n - 1/2| > 3/n) = 0$ for any $n$ (i.e. $C_n = 1$ for any $\delta \in (0, 1/2)$ and $n > \frac{3}{(1/2-\delta)}$) and a predictor $\Gamma$ such that $P_p^n(\mathrm{err}_n > 0) \leq 2^{1-n}$ for any $p \in [\delta, 1-\delta]$ and $\mathbf{P}(\mathrm{err}_n = 1) = 1$ for $n > 1$.*

*Proof.* Let $\mathbf{X} = \mathbf{Y} = \{0, 1\}$. We define the distributions $P_y$ as $P_y(X = y) = 1$, for each $y \in \mathbf{Y}$ (i.e. $\eta(x) = x$ for each $x$). The distribution $\mathbf{P}|_{\mathbf{Y}^\infty}$ is defined as a Markov distribution with transition probability matrix $\begin{pmatrix} 0 & 1 \\ 1 & 0 \end{pmatrix}$, i.e. it always generates sequences of labels $\ldots 01010101 \ldots$.

We define the predictor $\Gamma$ as follows

$$\Gamma_n := \begin{cases} 1 - x_n & \text{if } |\#\{i < n : y_i = 0\} - n/2| \leq 1, \\ x_n & \text{otherwise}. \end{cases}$$

So, in the case when the distribution $\mathbf{P}$ is used to generate the examples, $\Gamma$ is always seeing either $n-1$ zeros and $n$ ones, or $n$ zeros and $n$ ones which, consequently, will lead it to always predict the wrong label. It remains to note that this is almost improbable in the case of an i.i.d. distribution. □

Another point is the requirement on the frequencies of labels. In particular, the assumption (7) might appear redundant: if the rate of occurrence of some label tends to zero, can we just ignore this label without affecting the asymptotic? It appears that this is not the case, as the following example illustrates.

**Remark 2.** *There exist a distribution $\mathbf{P}$ on $\mathbf{X}^\infty$ which satisfies (1) and (2) but for which the nearest neighbour predictor is not consistent, i.e. the probability of error does not tend to zero.*

*Proof.* Let $\mathbf{X} = [0,1]$, let $\eta(x) = 0$ if $x$ is rational and $\eta(x) = 1$ otherwise. The distribution $P_1$ is uniform on the set of irrational numbers, while $P_0$ is any distribution such that $P(x) \ne 0$ for any rational $x$. (This construction is due to T. Cover.) The nearest neighbour predictor is consistent for any i.i.d. distribution which agrees with the definition, i.e. for any $p = P(Y = 1) \in [0,1]$.

Next we construct the distribution $\mathbf{P}|_{\mathbf{Y}^\infty}$. Fix some $\varepsilon$, $0 < \varepsilon < 1$. Assume that according to $\mathbf{P}$ the first label is always 1, (i.e. $\mathbf{P}(y_1 = 1) = 1$; the object is an irrational number). Next $k_1$ labels are always 0 (rationals), then follows 1, then $k_2$ zeros, and so on. It is easy to check that there exists such sequence $k_1, k_2, \ldots$ that with probability at least $\varepsilon$ we have

$$\max_{i<n:\ X_i \text{ is irrational}} P_1\{x : X_i \text{ is the nearest neighbour of } x\} \le \frac{1-\varepsilon}{m(n)},$$

where $m(n)$ is the total number of irrational objects up to the trial $n$. On each step $n$ such that $n = t + \sum_{j=1}^t k_t$ for some $t \in \mathbb{N}$ (i.e. on each irrational object) we have

$$\mathbf{E}(\text{err}_n(\Gamma, \mathbf{P}))$$
$$\ge \varepsilon \left( 1 - \sum_{j<n:\ X_j \text{ is irrational}} \mathbf{P}(X_j \text{ is the nearest neighbour of } X) \right) \ge \varepsilon^2$$

As irrational objects are generated infinitely often (that is, with intervals $k_i$), the probability of error does not tend to zero. □

Another question is whether the results can be generalised to the case of non-deterministically defined labels, which is often considered in literature. It should be noted that we consider the task of learning object-label dependence, ignoring the label-label dependence (and prohibiting any dependence apart from these). On one hand, it allows us to consider any sort of label-label dependence. On the other hand, the best bound on the probability of error we can obtain is the maximum of the class-conditional probabilities of error (as nothing is known about the probability of the next label), and not the so-called Bayes error, which is the best achievable bound in the i.i.d. case.

Thus, if we want to consider stochastically defined labels, we should restrict our attention to class-conditional probabilities of error. On this way also some obstacles can be met. In particular, the function $\eta$, which in this case is defined as $\eta(x) := \mathbf{P}(Y_n = 1 | X_n = x)$ should not depend on $n$, which will require more restrictive definition of constants $C_n$ and the condition (7). We leave this question for further investigation.

One more point which needs clarification is the choice of the constants $\varkappa_n$. We have fixed these constants for the sake of simplicity of notations, however, they can be made variable, as long as $\varkappa_n$ obeys the following condition.

$$\lim_{n \to \infty} \{n|p_n - p| \leq \varkappa_n\} = 0$$

almost surely for any $p \in (0,1)$ and any probability distribution $P$ on $\mathbf{X}$ such that $P(y=1) = p$, where $p_n := \frac{1}{n}\#\{i \leq n : Y_i = 0\}$.

# Appendix A: proofs for Section 2

Before proceeding with the proof of Theorem 1 we give some definitions and supplementary facts.

Define the conditional probabilities of error of $\Gamma$ as follows

$$\mathrm{err}_n^0(\Gamma, \mathbf{P}, z_0, \ldots, z_n) := \mathbf{P}(Y_{n+1} \neq \Gamma(z_1, \ldots, z_n, X_{n+1}) | Y_{n+1} = 0),$$

$$\mathrm{err}_n^1(\Gamma, \mathbf{P}, z_0, \ldots, z_n) := \mathbf{P}(Y_{n+1} \neq \Gamma(z_1, \ldots, z_n, X_{n+1}) | Y_{n+1} = 1),$$

(with the same notational convention as used with the definition of $\mathrm{err}_n(\Gamma)$). In words, for each $y \in \mathbf{Y} = \{0,1\}$ we define $\mathrm{err}_n^y$ as the probability of all $x \in \mathbf{X}$, such that $\Gamma$ makes an error on $n$'th trial, given that $Y_{n+1} = y$ and fixed $z_1, \ldots, z_n$.

For any $\mathbf{y} := (y_1, y_2, \ldots) \in \mathbf{Y}^\infty$, define $\mathbf{y}_n := (y_1, \ldots, y_n)$ and $p_n(\mathbf{y}) := \frac{1}{n}\#\{i \leq n : y_i = 0\}$, for $n > 1$.

Clearly (from the assumption (1)) the random variables $X_1, \ldots, X_n$ are mutually conditionally independent given $Y_1, \ldots, Y_n$, and by (2) they are distributed according to $P_{Y_i}$, $1 \leq i \leq n$. Hence, the following statement is valid.

**Lemma 1.** *Fix some $n > 1$ and some $\mathbf{y} \in \mathbf{Y}^\infty$ such that $\mathbf{P}((Y_1, \ldots, Y_{n+1}) = \mathbf{y}_{n+1}) \neq 0$. Then*

$$\mathbf{P}\big(\mathrm{err}_n^{y_{n+1}}(\Gamma) > \varepsilon \,\big|\, (Y_1, \ldots, Y_n) = \mathbf{y}_n\big)$$
$$= P_p^n\big(\mathrm{err}_n^{y_{n+1}}(\Gamma) > \varepsilon \,\big|\, (Y_1, \ldots, Y_n) = \mathbf{y}_n\big)$$

*for any $p \in (0,1)$.*

*Proof of Theorem 1.* Fix some $n > 1$, some $y \in \mathbf{Y}$ and such $\mathbf{y}^1 \in \mathbf{Y}^\infty$ that $n\delta \le p_n(\mathbf{y}^1) \le n(1-\delta)$ and $\mathbf{P}((Y_1, \ldots, Y_n) = \mathbf{y}_n^1) \ne 0$. Let $p := p_n(\mathbf{y}^1)/n$. We will find bounds on $\mathbf{P}\big(\mathrm{err}_n(\Gamma) > \varepsilon \mid (Y_1, \ldots, Y_n) = \mathbf{y}_n^1\big)$, first in terms of $\Delta$ and then in terms of $\bar{\Delta}$.

Lemma 1 allows us to pass to the i.i.d. case:

$$\mathbf{P}\big(\mathrm{err}_n^y(\Gamma, X_1, y_1^1, \ldots, X_n, y_n^1, X_{n+1}) > \varepsilon\big)$$
$$= P_p^n\big(\mathrm{err}_n^y(\Gamma, X_1, y_1^1, \ldots, X_n, y_n^1, X_{n+1}) > \varepsilon\big)$$

for any $y$ such that $\mathbf{P}(Y_1 = y_1^1, \ldots, Y_n = y_n^1, Y_{n+1} = y) \ne 0$ (recall that we use upper-case letters for random variables and lower-case for fixed variables, so that the probabilities in the above formula are labels-conditional).

Clearly, for $\delta \le p \le 1-\delta$ we have $\mathrm{err}_n(\Gamma, P_p) \le \max_{y \in \mathbf{Y}}(\mathrm{err}_n^y(\Gamma, P_p))$, and if $\mathrm{err}_n(\Gamma, P_p) < \varepsilon$ then $\mathrm{err}_n^y(\Gamma, P_p) < \varepsilon/\delta$ for each $y \in \mathbf{Y}$.

Let $m$ be such number that $m - \varkappa_m = n$. For any $\mathbf{y}^2 \in \mathbf{Y}^\infty$ such that $|mp_m(\mathbf{y}^2) - mp| \le \varkappa_m/2$ there exist such mapping $\pi : \{1, \ldots, n\} \to \{1, \ldots, m\}$ that $y_{\pi(i)}^2 = y_i^1$ for any $i \le n$. Define random variables $X_1' \ldots X_m'$ as follows: $X_{\pi(i)}' := X_i$ for $i \le n$, while the rest $\varkappa_m$ of $X_i'$ are some random variables independent from $X_1, \ldots, X_n$ and from each other, and distributed according to $P_p$ (a "ghost sample"). We have

$$P_p^n\big(\mathrm{err}_n^y(X_1, y_1^1, \ldots, X_n, y_n^1) > \varepsilon\big)$$
$$= P_p^m\Big(\mathrm{err}_n^y(X_1, y_1^1, \ldots, X_n, y_n^1) - \mathrm{err}_n^y(X_1', y_1^2, \ldots, X_m', y_m^2)$$
$$+ \mathrm{err}_n^y(X_1', y_1^2, \ldots, X_m', y_m^2) > \varepsilon\Big)$$
$$\le P_p^m\Big(\big|\mathrm{err}_n^y(X_1', y_1^2, \ldots, X_n', y_n^2) - \mathrm{err}_n^y(X_1, y_1^1, \ldots, X_n, y_n^1)\big| > \varepsilon/2\Big)$$
$$+ P_p^n\Big(\mathrm{err}_n^y(X_1', y_1^2, \ldots, X_n', y_n^2) > \varepsilon/2\Big).$$

Observe that $\mathbf{y}^2$ was chosen arbitrary (among sequences for which $|mp_m(\mathbf{y}^2) - mp| \le \varkappa_m/2$) and $(X_1, y_1^1, \ldots, X_n y_n^1)$ can be obtained from $(X_1', y_1^2, \ldots, X_m' y_m^2)$ by removing at most $\varkappa_m$ elements and applying some permutation. Thus the first term is bounded by

$$P_p^m\Big(\max_{j \le \varkappa_m;\ \pi:\{1,\ldots,m\} \to \{1,\ldots,m\}} |\mathrm{err}_m^y(\Gamma, Z_1, \ldots, Z_m) - \mathrm{err}_{m-j}^y(\Gamma, Z_{\pi(1)}, \ldots, Z_{\pi(m-j)})| > \varepsilon/2 \,\big|\, |mp(m) - mp| \le \varkappa_m/2\Big)$$
$$\le \frac{\Delta(P_p, m, \delta\varepsilon/2)}{P_p^n(|mp(m) - mp| \le \varkappa_m)} \le \frac{1}{1 - 1/\sqrt{m}} \Delta(P_p, m, \delta\varepsilon/2),$$

and the second term is bounded by $\frac{1}{1-1/\sqrt{m}}P_p^m(\mathrm{err}_m(\Gamma) > \delta\varepsilon/2)$. Hence

$$P_p^n\big(\mathrm{err}_n^y(X_1, y_1^1, \ldots, X_n, y_n^1) > \varepsilon\big)$$
$$\leq \alpha_n\big(\Delta(P_p, m, \delta\varepsilon/2) + P_p^m(\mathrm{err}_m(\Gamma) > \delta\varepsilon/2)\big). \quad (15)$$

Next we establish a similar bound in terms of $\bar\Delta$. For any $\mathbf{y}_n^2 \in \mathbf{Y}^n$ such that $|np_n(\mathbf{y}^2) - np| \leq \varkappa_n/2$ there exist such permutations $\pi_1, \pi_2$ of the set $\{1, \ldots, n\}$ that $y^1_{\pi_1(i)} = y^2_{\pi_2(i)}$ for any $i \leq n - \delta\varkappa_n$. Denote $n - \delta\varkappa_n$ by $n'$ and define random variables $X'_1 \ldots X'_n$ as follows: $X'_{\pi_2(i)} := X_{\pi_1(i)}$ for $i \leq n'$, while for $n' < i \leq n$ $X'_i$ are some "ghost" random variables independent from $X_1, \ldots, X_n$ and from each other, and distributed according to $P_p$. We have

$$P_p^n\big(\mathrm{err}_n^y(X_1, y_1^1, \ldots, X_n, y_n^1) > \varepsilon\big)$$
$$\leq P_p^{n+\varkappa_n}\Big(\big|\mathrm{err}_n^y(X'_1, y_1^2, \ldots, X'_n, y_n^2) - \mathrm{err}_n^y(X_1, y_1^1, \ldots, X_n, y_n^1)\big| > \varepsilon/2\Big)$$
$$+ P_p^n\Big(\mathrm{err}_n^y(X'_1, y_1^2, \ldots, X'_n, y_n^2) > \varepsilon/2\Big),$$

Again, as $\mathbf{y}^2$ was chosen arbitrary (among sequences for which $|np_n(\mathbf{y}^2) - np| \leq \varkappa_n/2$) and $(X_1, y_1^1, \ldots, X_n y_n^1)$ differs from $(X'_1, y_1^2, \ldots, X'_n y_n^2)$ in at most $\varkappa_n$ elements, up to some permutation. Thus the first term is bounded by

$$P_p^n\Big(\sup_{j<\varkappa_n;\pi:\{1,\ldots,n\}\to\{1,\ldots,n\};z'_{n-j},\ldots,z'_n} |\mathrm{err}_n^y(Z_1, \ldots, Z_n)$$
$$- \mathrm{err}_n^y(\zeta_1, \ldots, \zeta_n)| > \varepsilon/2 \,\big|\, |np(n) - np| \leq \varkappa_n/2\Big)$$
$$\leq \alpha_n\bar\Delta(P_p, n, \delta\varepsilon/2),$$

and the second term is bounded by $\alpha_n P_p^n(\mathrm{err}_n(\Gamma) > \delta\varepsilon/2)$. Hence

$$P_p^n\big(\mathrm{err}_n^y(X_1, y_1^1, \ldots, X_n, y_n^1) > \varepsilon\big)$$
$$\leq \alpha_n\big(\bar\Delta(P_p, n, \delta\varepsilon/2) + P_p^n(\mathrm{err}_n(\Gamma) > \delta\varepsilon/2)\big). \quad (16)$$

Finally, as $\mathbf{y}^1$ was chosen arbitrary among sequences $\mathbf{y} \in \mathbf{Y}^\infty$ such that $n\delta \leq p_n(\mathbf{y}^1) \leq n(1-\delta)$ from (15) and (16) we obtain (5) and (6). $\square$

## Appendix B: proofs for Section 3

The first part of the proof is common for theorems 2 and 3. Let us fix some distribution $\mathbf{P}$ satisfying conditions of the theorems. It is enough to show

that
$$\sup_{p \in [\delta, 1-\delta]} E^\infty(\mathrm{err}_n(\Gamma, P_p, Z_1, \ldots, Z_n)) \to 0$$
and
$$\sup_{p \in [\delta, 1-\delta]} E^\infty(\bar{\Delta}(P_p, n, Z_1, \ldots, Z_n)) \to 0$$
for nearest neighbour and partitioning predictor, and apply Corollary 1.

Observe that both predictors are symmetric, i.e. do not depend on the order of $Z_1, \ldots, Z_n$. Thus, for any $z_1, \ldots, z_n$

$$\bar{\Delta}(P_p, n, z_1, \ldots, z_n) = \sup_{j \leq \varkappa_n;\ \pi:\{1,\ldots,n\} \to \{1,\ldots,n\}, z'_{n-j}, \ldots, z'_n}$$
$$|\mathrm{err}_n(\Gamma, P_p, z_1, \ldots, z_n) - \mathrm{err}_n(\Gamma, P_p, z_{\pi(1)}, \ldots, z_{\pi(n-j)}, z'_{n-j}, \ldots, z'_n)|,$$

where the maximum is taken over all $z'_i$ consistent with $\eta$, $n - j \leq i \leq n$. Define also the class-conditional versions of $\bar{\Delta}$:

$$\bar{\Delta}^y(P_p, n, z_1, \ldots, z_n) := \sup_{j \leq \varkappa_n;\ \pi:\{1,\ldots,n\} \to \{1,\ldots,n\}, z'_{n-j}, \ldots, z'_n}$$
$$|\mathrm{err}_n^y(\Gamma, P_p, z_1, \ldots, z_n) - \mathrm{err}_n^y(\Gamma, P_p, z_{\pi(1)}, \ldots, z_{\pi(n-j)}, z'_{n-j}, \ldots, z'_n)|.$$

Note that (omitting $z_1, \ldots, z_n$ from the notation) $\mathrm{err}_n(\Gamma, P_p) \leq \mathrm{err}_n^0(\Gamma, P_p) + \mathrm{err}_n^1(\Gamma, P_p)$ and $\bar{\Delta}(P_p, n) \leq \bar{\Delta}^0(P_p, n) + \bar{\Delta}^1(P_p, n)$. Thus, it is enough to show that
$$\sup_{p \in [\delta, 1-\delta]} E^\infty(\mathrm{err}_n^1(\Gamma, P_p)) \to 0 \tag{17}$$
and
$$\sup_{p \in [\delta, 1-\delta]} E^\infty(\bar{\Delta}^1(P_p, n)) \to 0. \tag{18}$$

Observe that for each of the predictors in question the probability of error given that the true label is 1 will not decrease if an arbitrary (possibly large) portion of training examples labelled with ones is replaced with an arbitrary (but consistent with $\eta$) portion of the same size of examples labelled with zeros. Thus, for any $n$ and any $p \in [\delta, 1-\delta]$ we can decrease the number of ones in our sample (by replacing the corresponding examples with examples from the other class) down to (say) $\delta/2$, not decreasing the probability of error on examples labelled with 1. So,

$$E^\infty(\mathrm{err}_n^1(\Gamma, P_p)) \leq E^\infty(\mathrm{err}_n^1(\Gamma, P_{\delta/2}|p_n = \delta/2)) + P_p(p_n \leq \delta/2), \tag{19}$$

where as usual $p_n := \frac{1}{n}\#\{i \leq n : y_i = 1\}$. Obviously, the last term (quickly) tends to zero. Moreover, it is easy to see that

$$
\begin{aligned}
E^\infty(\mathrm{err}_n^1(\Gamma, P_{\delta/2})|p_n &= n(\delta/2)) \\
&\leq E^\infty\big(\mathrm{err}_n^1(\Gamma, P_{\delta/2})\big||n(\delta/2) - p_n| \leq \varkappa_n/2\big) + E^\infty(\bar{\Delta}^1(P_{\delta/2}, n)) \\
&\leq \frac{1}{1 - 1/\sqrt{n}} E^\infty(\mathrm{err}_n^1(\Gamma, P_{\delta/2})) + E^\infty(\bar{\Delta}^1(P_{\delta/2}, n)). \quad (20)
\end{aligned}
$$

The first term tends to zero, as it is known from the results for i.i.d. processes; thus, to establish (17) we have to show that

$$E(\bar{\Delta}^1(P_p, n, Z_1, \ldots, Z_n)) \to 0 \quad (21)$$

for any $p \in (0, 1)$.

We will also show that (21) is sufficient to prove (18). Indeed,

$$
\bar{\Delta}^1(P_p, n, z_1, \ldots, z_n) \leq \mathrm{err}_n^1(\Gamma, P_p, z_1, \ldots, z_n) + \\
\sup_{j \leq \varkappa_n;\ \pi:\{1,\ldots,n\} \to \{1,\ldots,n\}, z'_{n-j},\ldots,z'_n} \mathrm{err}_n^1(\Gamma, P_p, z_{\pi(1)}, \ldots, z_{\pi(n-j)}, z'_{n-j}, \ldots, z'_n)
$$

Denote the last summand by $D$. Again, we observe that $D$ will not decrease if an arbitrary (possibly large) portion of training examples labelled with ones is replaced with an arbitrary (but consistent with $\eta$) portion of the same size of examples labelled with zeros. Introduce $\widetilde{\Delta}^1(P_p, n, z_1, \ldots, z_n)$ as $\bar{\Delta}^1(P_p, n, z_1, \ldots, z_n)$ with $\varkappa_n$ in the definition replaced by $\frac{2}{\delta}\varkappa_n$. Using the same argument as in (19) and (20) we have

$$E^\infty(D) \leq \frac{1}{1 - 1/\sqrt{n}}\big(E^\infty(\widetilde{\Delta}^1(P_{\delta/2}, n)) + E^\infty(\mathrm{err}_n(\Gamma, P_{\delta/2}))\big) + P_p(p_n \leq \delta/2).$$

Thus, (18) holds true if (21) and

$$E^\infty(\widetilde{\Delta}^1(P_p, n, Z_1, \ldots, Z_n)) \to 0. \quad (22)$$

Finally, we will prove (21); it will be seen that the proof of (22) is analogous (i.e. replacing $\varkappa_n$ by $\frac{2}{\delta}\varkappa_n$ does not affect the proof). Note that

$$
E^\infty(\bar{\Delta}(P_p, n, Z_1, \ldots, Z_n)) \leq P_p\bigg(\sup_{j \leq \varkappa_n;\ \pi:\{1,\ldots,n\} \to \{1,\ldots,n\}, z'_{n-j},\ldots,z'_n}
$$

$$\big|\mathrm{err}_n(\Gamma, P_p, Z_1, \ldots, Z_n) \neq \mathrm{err}_n(\Gamma, P_p, Z_{\pi(1)}, \ldots, Z_{\pi(n-j)}, z'_{n-j}, \ldots, z')\big|\bigg),$$

where the maximum is taken over all $z'_i$ consistent with $\eta$, $n - j \leq i \leq n$. The last expression should be shown to tend to zero. This we will prove for each of the predictors separately.

*Nearest Neighbour predictor.* Fix some distribution $P_p$, $0 < p < 1$ and some $\varepsilon > 0$. Fix also some $n \in \mathbb{N}$ and define (leaving $x_1, \ldots, x_n$ implicit)

$$B_n(x) := P_p^{n+1}\{t \in \mathbf{X} : t \text{ and } x \text{ have the same nearest neighbour among } x_1, \ldots, x_n\}$$

and $B_n := E(B_n(X))$ Note that $E^\infty(B_n) = 1/n$, where the expectation is taken over $X_1, \ldots, X_n$. Define $\mathcal{B} := \{(x_1, \ldots, x_n) \in \mathbf{X}^n : B_n \leq 1/n\varepsilon\}$ and $\mathcal{A}(x_1, \ldots, x_n) := \{x : B_n(x) \leq 1/n\varepsilon^2\}$. Applying Markov's inequality twice, we obtain

$$E^\infty(\bar{\Delta}(P_p, n)) \leq E^\infty(\bar{\Delta}(P_p, n)|(X_1, \ldots, X_n) \in \mathcal{B}) + \varepsilon$$
$$\leq E^\infty \Big( \sup_{j \leq \varkappa_n;\ \pi:\{1,\ldots,n\}\to\{1,\ldots,n\}, z'_{n-j},\ldots,z'_n}$$
$$P_p\{x : \mathrm{err}_n(\Gamma, P_p, Z_1, \ldots, Z_n) \neq \mathrm{err}_n(\Gamma, P_p, Z_{\pi(1)}, \ldots, Z_{\pi(n-j)}, z'_{n-j}, \ldots, z'_n)$$
$$\Big| x \in \mathcal{A}(X_1, \ldots, X_n)\} \Big| (X_1, \ldots, X_n) \in \mathcal{B} \Big) + 2\varepsilon. \tag{23}$$

Removing one point $x_i$ from a sample $x_1, \ldots, x_n$ we can only change the value of $\Gamma$ in the area

$$\{x \in \mathbf{X} : x_i \text{ is the nearest neighbour of x}\} = B_n(x_i),$$

while adding one point $x_0$ to the sample we can change the value of $\Gamma$ in the area

$$D_n(x_0) := \{x \in \mathbf{X} : x_0 \text{ is the nearest neighbour of x}\}.$$

It can be shown that the number of examples (among $x_1, \ldots, x_n$) for which a point $x_0$ is the nearest neighbour is not greater than a constant $\gamma$ which depends only the space $\mathbf{X}$ (see [10], Corollary 11.1). Thus,

$$D_n(x_0) \subset \cup_{i=j_1, \ldots, j_\gamma} B_n(x_i)$$

for some $j_1, \ldots, j_\gamma$, and so

$$E^\infty(\bar{\Delta}(P_p, n)) \leq 2\varepsilon + 2(\gamma + 1)\varkappa_n E^\infty(\max_{x \in \mathcal{A}(X_1, \ldots, X_n)} B_n(x)|(X_1, \ldots, X_n) \in \mathcal{B})$$
$$\leq 2\varkappa_n \frac{\gamma + 1}{n\varepsilon^2} + 2\varepsilon,$$

which, increasing $n$, can be made less than $3\varepsilon$. $\square$

*Partitioning predictor.* For any measurable sets $\mathcal{B} \subset \mathbf{X}^n$ and $\mathcal{A} \subset \mathbf{X}$ define

$$D(\mathcal{B}, \mathcal{A}) := E^\infty \Big( \sup_{j \leq \varkappa_n;\ \pi:\{1,\ldots,n\}\to\{1,\ldots,n\}, z'_{n-j},\ldots,z'_n}$$

$$P_p\{x : \mathrm{err}_n(\Gamma, P_p, Z_1, \ldots, Z_n) \neq \mathrm{err}_n(\Gamma, P_p, Z_{\pi(1)}, \ldots, Z_{\pi(n-j)}, z'_{n-j}, \ldots, z'_n)$$

$$|x \in \mathcal{A}\}\big|(X_1, \ldots, X_n) \in \mathcal{B}\Big) + 2\varepsilon.$$

and $D := D(\mathbf{X}^n, \mathbf{X})$.

Fix some distribution $P_p$, $0 < p < 1$ and some $\varepsilon > 0$. Introduce

$$\hat\eta(x, X_1, \ldots, X_n) := \frac{1}{N(x)} \sum_{i=1}^n I_{Y_i=1} I_{X_i \in A(x)}$$

$(X_1, \ldots X_n$ will usually be omitted). From the consistency results for i.i.d. model (see, e.g. [10], Theorem 6.1) we know that $E^{n+1}|\hat\eta_n(X) - \eta(X)| \to 0$ (the upper index in $E^{n+1}$ indicating the number of examples it is taken over).

Thus, $E|\hat\eta_n(X) - \eta(X)| \leq \varepsilon^4$ from some $n$ on. Fix any such $n$ and let $\mathcal{B} := \{(x_1, \ldots, x_n) : E|\hat\eta_n(X) - \eta(X)| \leq \varepsilon^2\}$. By Markov inequality we obtain $P_p(\mathcal{B}) \geq 1 - \varepsilon^2$. For any $(x_1, \ldots, x_n) \in \mathcal{B}$ let $\mathcal{A}(x_1, \ldots, x_n)$ be the union of all cells $A_i^n$ for which $E(|\hat\eta_n(X) - \eta(X)||X \in A_i^n) \leq \varepsilon$. Clearly, with $x_1, \ldots, x_n$ fixed, $P_p(X \in \mathcal{A}(x_1, \ldots, x_n)) \geq 1 - \varepsilon$. Moreover, $D \leq D(\mathcal{B}, \mathcal{A}) + \varepsilon + \varepsilon^2$.

Fix $\mathcal{A} := (x_1, \ldots, x_n)$ for some $(x_1, \ldots, x_n) \in \mathcal{B}$. Since $\eta(x)$ is always either 0 or 1, to change a decision in any cell $A \subset \mathcal{A}$ we need to add or remove at least $(1-\varepsilon)N(A)$ examples, where $N(A) := N(x)$ for any $x \in A$. Let $N(n) := E(N(X))$ and $A(n) := E(P_p(A(X))$. Clearly, $\frac{N(n)}{nA(n)} = 1$ for any $n$, as $E\frac{N(X)}{n} = A(n)$.

As before, using Markov inequality and shrinking $\mathcal{A}$ if necessary we can have $P_p(\frac{\varepsilon^2 n A(X)}{N(n)} \leq \varepsilon | X \in \mathcal{A}) = 1$, $P_p(\frac{\varepsilon^2 n A(n)}{N(X)} \leq \varepsilon | X \in \mathcal{A}) = 1$, and $D \leq D(\mathcal{B}, \mathcal{A}) + 3\varepsilon + \varepsilon^2$. Thus, for all cells $A \subset \mathcal{A}$ we have $N(A) \geq \varepsilon n A(n)$, so that the probability of error can be changed in at most $2\frac{\varkappa_n}{(1-\varepsilon)\varepsilon n A(n)}$ cells; but the probability of each cell is not greater than $\frac{N(n)}{\varepsilon n}$. Hence $E^\infty(\bar\Delta(P_p, n)) \leq 2\frac{\varkappa_n}{n(1-\varepsilon)\varepsilon^2} + 3\varepsilon + \varepsilon^2$. $\square$

## Appendix C: proofs for Section 4

*Proof of Theorem 4.* Fix some probability distribution $P_p$ and some $n \in \mathbb{N}$. Let $\varphi^\times$ be any decision rule $\varphi \in \mathcal{C}$ picked by $\Gamma_{n-\varkappa_n}$ on which (along with the corresponding permutation) the maximum

$$\max_{j \leq \varkappa_n;\ \pi:\{1,\ldots,n\}\to\{1,\ldots,n\}} |\operatorname{err}_n(\Gamma, z_1, \ldots, z_n) - \operatorname{err}_{n-j}(\Gamma, z_{\pi(1)}, \ldots, z_{\pi(n-j)})|$$

is reached. We need to estimate $P^n(|\operatorname{err}(\varphi^*) - \operatorname{err}(\varphi^\times)| > \varepsilon)$.

Clearly, $|\overline{\operatorname{err}}_n(\varphi^\times) - \overline{\operatorname{err}}_n(\varphi^*)| \leq \varkappa_n$, as $\varkappa_n$ is the maximal number of errors which can be made on the difference of the two samples.

Moreover,

$$P^n(|\operatorname{err}(\varphi_n^*) - \operatorname{err}(\varphi^\times)| > \varepsilon)$$
$$\leq P^n(|\operatorname{err}(\varphi_n^*) - \frac{1}{n}\overline{\operatorname{err}}_n(\varphi^*)| > \varepsilon/2)$$
$$+ P^n(|\frac{1}{n}\overline{\operatorname{err}}_n(\varphi^\times) - \operatorname{err}(\varphi^\times)| > \varepsilon/2 - \varkappa_n/n)$$

Observe that

$$P^n(\sup_{\varphi \in \mathcal{C}} |\frac{1}{n}\overline{\operatorname{err}}_n(\varphi) - \operatorname{err}(\varphi)| > \varepsilon) \leq 8\mathcal{S}(\mathcal{C}, n)e^{-n\varepsilon^2/32}, \tag{24}$$

see [10], Theorem 12.6. Thus,

$$\Delta(P_p, n, \varepsilon) \leq 16\mathcal{S}(\mathcal{C}, n)e^{-n(\varepsilon/2 - \varkappa_n/n)^2/32} \leq 16\mathcal{S}(\mathcal{C}, n)e^{-n\varepsilon^2/512}$$

for $n > 4/\varepsilon^2$. So,

$$\mathbf{P}(\operatorname{err}_n(\Gamma, \mathbf{P}) > \varepsilon) \leq I_{\sup_{p \in [\delta, 1-\delta]} \operatorname{err}(\varphi_{P_p}, P_p) > \varepsilon/2}$$
$$+ 16\alpha C_n^{-1}\mathcal{S}(\mathcal{C}, n)e^{-n\delta^2\varepsilon^2/2048} + (1 - C_n).$$

It remains to notice that

$$\operatorname{err}(\varphi_{P_p}, P_p) = \inf_{\varphi \in \mathcal{C}}(p\operatorname{err}^1(\varphi, P_p) + (1-p)\operatorname{err}^0(\varphi, P_p))$$
$$\leq \inf_{\varphi \in \mathcal{C}}(\operatorname{err}^1(\varphi, P_{1/2}) + \operatorname{err}^0(\varphi, P_{1/2})) = 2\operatorname{err}(\varphi_{P_{1/2}}, P_{1/2})$$

for any $p \in [0, 1]$.

So far we have proven (11) and (12); (13) and (14) can be proven analogously, only for the case $\eta \in \mathcal{C}$ we have

$$P^n(\sup_{\varphi \in \mathcal{C}} |\frac{1}{n}\overline{\operatorname{err}}_n(\varphi) - \operatorname{err}(\varphi)| > \varepsilon) \leq \mathcal{S}(\mathcal{C}, n)e^{-n\varepsilon}$$

instead of (24), and $\operatorname{err}(\varphi_{P_p}, P_p) = 0$. $\square$